\pdfoutput=1

\documentclass[11pt,table,dvipsnames]{article}

\usepackage[final]{acl}
\usepackage{times}
\usepackage{latexsym}

\usepackage[T1]{fontenc}

\usepackage[utf8]{inputenc}

\usepackage{microtype}

\usepackage{inconsolata}
\usepackage{microtype}
\usepackage{latexsym}
\usepackage{dsfont}
\usepackage{booktabs} 
\usepackage{CJKutf8}
\usepackage{graphicx}
\usepackage{bigstrut}
\usepackage{multirow}
\usepackage{amsmath}
\usepackage{multirow, makecell, caption}
\usepackage{floatrow}
\newfloatcommand{capbtabbox}{table}[][\FBwidth]
\usepackage{xcolor}
\usepackage[normalem]{ulem}
\usepackage{amssymb}
\usepackage{amsfonts}
\usepackage{multicol}
\usepackage{tipa}
\usepackage{amsmath}
\usepackage{bm}
\usepackage[ruled,linesnumbered]{algorithm2e}
\usepackage{tikz}
\usepackage{paralist}
\usepackage{IEEEtrantools}
\usepackage[switch]{lineno}
\usepackage{enumitem}
\usepackage{multirow, makecell, caption}
\usepackage{arydshln}
\usepackage{verbatim}
\usepackage[normalem]{ulem}
\usepackage{amssymb}
\usepackage{amsfonts}
\usepackage{multicol}
\usepackage{amssymb}
\usepackage{pifont}
\usepackage{csquotes}
\usepackage{xspace}
\usepackage{paralist}
\usepackage{mdwlist}
\usepackage{subcaption}
\usepackage{makecell}
\usepackage{array}
\usepackage{bm}
\usepackage{colortbl}
\usepackage{dsfont}
\usepackage{url}
\usepackage{booktabs} 
\usepackage{graphicx}
\usepackage{tabularx}
\usepackage{amsmath}
\usepackage{bbm}
\usepackage{pgfplots}
\usepackage{soul} 
\usepackage{listings}
\usepackage[tableposition=top]{caption}
\usepackage{color,xcolor} 
\usepackage{algorithmic}
\newcommand{\eg}{{e.g.}}
\newcommand{\ie}{{i.e.}} 

\newcommand{\method}{\textsc{EvoAgent}\xspace}\makeatletter
\def\adl@drawiv#1#2#3{%
        \hskip.5\tabcolsep
        \xleaders#3{#2.5\@tempdimb #1{1}#2.5\@tempdimb}%
                #2\z@ plus1fil minus1fil\relax
        \hskip.5\tabcolsep}
\newcommand{\cdashlinelr}[1]{%
  \noalign{\vskip\aboverulesep
           \global\let\@dashdrawstore\adl@draw
           \global\let\adl@draw\adl@drawiv}
  \cdashline{#1}
  \noalign{\global\let\adl@draw\@dashdrawstore
           \vskip\belowrulesep}}
\makeatother

\title{\method: Towards Automatic Multi-Agent Generation via Evolutionary Algorithms}

\author{%
Siyu Yuan$^1$\thanks{~~The first two authors have equal contributions. This work was done when the first author was an intern at
Microsoft Research Asia.},
Kaitao Song$^2$\footnotemark[1] \thanks{~~Corresponding authors.},\\ 
\bf Jiangjie Chen$^1$,
Xu Tan$^2$,
Dongsheng Li$^2$,
Deqing Yang$^1$\footnotemark[2]\\
Fudan University$^1$,
Microsoft Research Asia$^2$\\
\texttt{syyuan21@m.fudan.edu.cn},
 \texttt{\{kaitaosong, xuta, dongsli\}@microsoft.com}\\
\texttt{\{jjchen19,yangdeqing\}@fudan.edu.cn}\\
\url{https://evo-agent.github.io}
}

\begin{document}

\maketitle

\begin{abstract}
The rise of powerful large language models (LLMs) has spurred a new trend in building LLM-based autonomous agents for solving complex tasks, especially multi-agent systems.
Despite the remarkable progress, we notice that existing works are heavily dependent on human-designed frameworks, which greatly limits the functional scope and scalability of agent systems.
How to automatically extend the specialized agent to multi-agent systems to improve task-solving capability still remains a significant challenge. 
In this paper, we introduce \method, a generic method to automatically extend specialized agents to multi-agent systems via the evolutionary algorithm, thereby improving the effectiveness of LLM-based agents in solving tasks.
Specifically, we consider the existing agent frameworks as the initial individual and then apply a series of evolutionary operators (\eg, mutation, crossover, selection, etc.) to generate multiple agents with diverse settings. 
Experimental results across various tasks show that \method can significantly enhance the task-solving capability of LLM-based agents, and can be generalized to any LLM-based agent framework to extend them into multi-agent systems.
Resources are available at \url{https://evo-agent.github.io/}.

\end{abstract}

\section{Introduction}
\label{sec:intro}
Recently, the advent of large language models (LLMs)~\cite{OpenAI2023GPT4, geminiteam2023gemini, Hugo2023LLaMa2, Anthropic2024Claude3} have shown remarkable capabilities in solving language understanding, reasoning, and generation tasks. 
Based on the foundation of LLMs, many research works~\cite{gravitas2023auto, Yongliang2023HuggingGPT, Yohei2023BabyAGI, Timo2023Toolformer, Jason2022CoT, hong2024metagpt, Park2023GenerativeAgents} have discovered that by empowering multiple advanced skills (\eg, planning, tool, memory and so on), we can develop more powerful autonomous agents to solve more challenging tasks. 
Therefore, how to design and leverage LLM-based autonomous agents to tackle more diverse and complex real-world applications has attracted enormous interest.

\begin{figure}[t]
    \centering
    \includegraphics[width=1.0\linewidth]{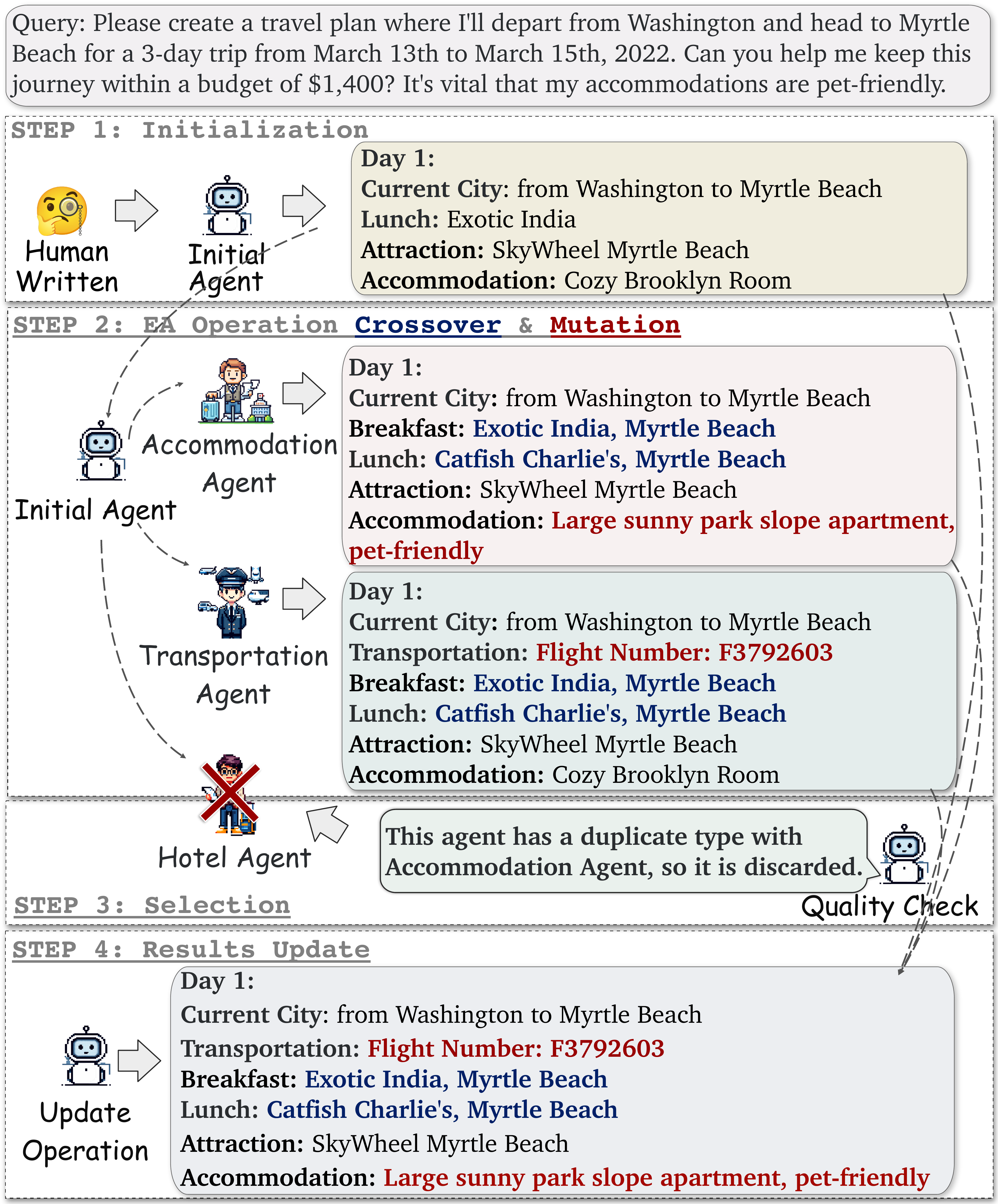}
    \caption{
    The illustration of \method. With the generated multiple specialized agents, \method can generate a better travel plan to meet user preferences.
    For EA operators, {\color[rgb]{0.2,0.3,0.7}{\textbf{Crossover}}} can improve the results of parent agents by adjusting existing details (\eg, the information marked as {\color[rgb]{0.2,0.3,0.7}{\textbf{blue}}}). 
{\color[rgb]{0.7,0,0.1}{\textbf{Mutation}}} can introduce new variations to refine the results of parent agents by generating child agents with new characteristics (\eg, the information marked as {\color[rgb]{0.7,0,0.1}{\textbf{red}}}). 
}
    \label{figue:evoagent}
\end{figure}

Generally, many real-world scenarios are usually complex, encompassing a variety of challenging tasks that are beyond the capability of a single agent. 
To address this point, we notice that human society is composed of vast individuals, each possessing their unique characteristics. 
By selecting, orchestrating, and cooperating with different individuals, humans can form an efficient team group to handle complicated missions in the real world. 
Therefore, there has been an increasing trend to develop multi-agent collaboration frameworks~\cite{Park2023GenerativeAgents,li2023camel,wu2023autogen,hong2024metagpt} to simulate human behaviors for solving complex tasks.
By developing a series of specialized agents with diverse settings, multi-agent systems enable us to reveal emergent abilities among multiple agents and synergize their specialized expertise to achieve superior performance, akin to simulating human populations. 
Nevertheless, it is worthy noting that, in most of (multi)-agent frameworks, their designs heavily depend on handcrafted settings, including character roles, task scopes, skills, and prompt settings. 
Although we admit that meticulous human design is quite useful for instructing LLM-based agents to understand tasks, it also limits scaling up the number of agents to further improve performance due to expensive human labor.
Considering the increasing popularity of LLM-based autonomous agents, how to create a generic agent generation paradigm to automatically build multi-agent systems is a critical challenge.

In this paper, we introduce a novel method, \method, that formulates agent generation as the evolutionary processing~\cite{Thomas1993Evolutionary} in human society. 
Specifically, to align human society, each agent can be considered as individuals that can procreate its population across successive generations. 
Motivated by this mechanism, we can simulate such a human behavior to automatically generate multiple agents based on any pre-defined agents.
Therefore, as shown in Figure~\ref{figue:evoagent}, \method can be considered as a one-shot agent generation method that starts from a specialized agent as the initial agent, and then considers its settings (\eg, role, skills, prompts, and so on) as the variables to be evolved. 
With a series operation of EAs (\eg, selection, crossover, mutation), \method can automatically create multiple evolutionary agents based on the initial specialized agent. 
Moreover, \method is not limited to the infrastructure of agent frameworks, as it is a generic multi-agent generation method.
Thus, it can be applied to any agent framework and expanded to multi-agent systems without any extra human effort.  

We conduct experiments on multiple datasets, including knowledge-based question answering and multi-modal reasoning ($\mathsection$~\ref{sec:nlp_task}), interactive scientific solving ($\mathsection$~\ref{sec:scienceworld}) and real-world complex planning ($\mathsection$~\ref{sec:travelplanner}). 
Experimental results indicate that \method can generate multiple agents with diverse skills and harness their capabilities to consistently improve the performance of the model in different scenarios. 
Besides, to validate the scalability of \method in creating massive agents, we also apply our method to some conversational scenarios (\eg, debate), and the results also indicate the potential of \method in generating multiple diverse agents.
Overall, the contributions of this paper can be summarized as below:
\begin{itemize}[leftmargin=*]
    \item We introduce \method, a simple and generic multi-agent generation method to improve the effectiveness of LLM-based agents in solving tasks. \method can automatically generate new specialized agents and is applicable to any agent framework.
    \item We formulate the agent generation processing as an evolutionary pipeline, that encompasses multiple operators (\eg, selection, crossover, mutation) to generate agent population without additional human supervision.    
    \item We conduct extensive experiments on various tasks and demonstrate the effectiveness, scalability, and generality of our \method. Particularly, \method can significantly enhance the performance of LLM-based agents in both challenging open-world scenarios and complex real-world planning by generating more specialized agents.
\end{itemize}

\section{Related Work}
\label{sec:related}
\paragraph{LLM-based Autonomous Agents}
With the emergence of powerful large language models~\citep{OpenAI2023GPT4,geminiteam2023gemini,Hugo2023LLaMa2,Anthropic2024Claude3}, many researchers have endeavored to develop advanced autonomous agents~\cite{gravitas2023auto,Yongliang2023HuggingGPT,Yohei2023BabyAGI} empowered by multiple high-level LLM skills (\eg, personas~\cite{Park2023GenerativeAgents,wang2024incharacter,chen2024persona}, planning~\cite{Jason2022CoT,chen2023put,zhang2024timearena,yuan2023distilling}, tool~\cite{Timo2023Toolformer, Yongliang2023HuggingGPT,shen2023taskbench,yuan2024easytool} and memory~\cite{Jason2015Memory,shinn2023reflexion}). Some of them also extend agent frameworks to multi-agent collaboration (\eg, MetaGPT~\cite{li2023metaagents}, Generative Agents~\cite{Park2023GenerativeAgents}, AutoGen~\cite{wu2023autogen}, Camel~\cite{li2023camel}, AgentVerse~\cite{chen2024agentverse} and so on), by designing multiple specific roles. These systems also demonstrate satisfactory performance in addressing massive, challenging tasks. However, it is worth noting that most of the popular agent frameworks heavily relied on handcrafted designs. The abundant human efforts necessitated by these systems also limit the adaptability and flexibility of agents to handle unexpected challenges~\citep{qian2023communicative,he-etal-2023-lego,chen2024agentverse,hong2024metagpt}. In this paper, we propose \method, a method that can be applied to any LLM-based agent framework and easily extend to multi-agent systems. 
By using EA, our method allows us to iteratively generate and optimize multiple agents with diverse settings.

\paragraph{Agent Generation}
Recent studies have shown that assigning personas or roles to LLM-based autonomous agents can influence their behavior and performance in generation tasks~\citep{xu2023expertprompting,deshpande-etal-2023-toxicity,10.1145/3586183.3606763,li2023camel}. Current methods primarily involve manually assigning these personas and limit multi-agent collaboration to single or fixed roles, which requires significant human effort and hinders generalization~\citep{li2023camel,wu2023autogen,li2023metaagents,hong2024metagpt}. 
To address this, some frameworks like AgentVerse~\citep{chen2024agentverse} and AutoAgents~\citep{chen2023autoagents} have been proposed to automatically generate unlimited agents for collaborative task completion. 
However, these methods still heavily depend on human-designed interventions, which limits their scalability and functionality. 
For example, AutoAgents requires agent settings to satisfy a ``Planner - Agent Observer - Plan Observer'' framework. 
Meanwhile, AgentVerse formulates a pipeline of ``Expert Recruitment - Collaborative Decision Making - Action Execution - Evaluation'' to build agents. These architectures also limit the task scope of designing agents. 
In contrast, \method can automatically formulate the current agent frameworks to multi-agent systems with high-quality generated specialized agents by using EAs, which is flexible and adaptable to various agent frameworks.

\section{Method}
\label{sec:method}

Generally, human society comprises a broad spectrum of individuals from diverse cultures, encompassing multiple generations. 
To solve specific tasks, human society usually involves a lot of expert individuals and aggregates their specialized expertise to achieve better answer. 
Thus, it can be considered as the foundation to facilitate multi-agent collaborations. 
To fulfill this point, how to automatically create multiple agents would be very critical. 
Inspired by evolutionism, we formulate agent generation as an evolutionary process to generate multiple agents without any human labor.

\begin{algorithm*}[t]
\caption{Multi-Agent Generation with Evolutionary Algorithm}
\label{alg:evoagent}

\SetKwInput{KwRequire}{Require}
\SetKwInput{KwInput}{Input}
\SetKwInput{KwOutput}{Output}

\KwRequire{Initial agent $A_{(0,0)}$, population size $N$ per iteration, number of iterations $T$, quality-check module $\mathbf{LLM}_{\texttt{Quality}}(\cdot)$, evolutionary operations $\mathbf{Evo}_{\texttt{Crossover}}(\cdot)$ and $\mathbf{Evo}_{\texttt{Mutation}}(\cdot)$,
$\mathbf{Evo}_{\texttt{Update}}(\cdot)$
}
\KwInput{Initial result $R_0$ derived from $A_{(0,0)}$}
\KwOutput{Final result $R_T$}

\For{$t = 1$ \KwTo $T$}{
    \textbf{Crossover:}
    Update the settings of parent agents based on their generated results and initial agent:
    $\{A^{'}_{(0,t-1)}, A^{'}_{(1,t-1)}, ..., A^{'}_{(N-1,t-1)}\} \leftarrow \mathbf{Evo}_{\texttt{Crossover}}(\{R_{(0,t-1)}, R_{(1,t-1)}, ..., R_{(N-1,t-1)}\}, A_{(0,0)})$\;

    
    \textbf{Mutation:}
    Generate $N' (N' > N) $ child agents based on parent agents and initial agent:
    $\{A_{(0,t)}, A_{(1,t)}, ..., A_{(N'-1,t)}\} \leftarrow \mathbf{Evo}_{\texttt{Mutation}}(\{A^{'}_{(0,t-1)}, A^{'}_{(1,t-1)}, ..., A^{'}_{(N-1,t-1)}\}, A_{(0,0)})$\
    
    

    \textbf{Selection}:
    Select high-quality agents with quality-check module: $\{A_{(0,t)}, A_{(1,t)}, ..., A_{(N-1,t)}\} \leftarrow \mathbf{LLM}_{\texttt{Quality}}(\{A_{(0,t)}, A_{(1,t)}, ..., A_{(N'-1,t)}\}, \{A_{(N,t')}\}_{t'=1}^{t'=t-1}$)\;
    

    \textbf{Result Update:}
    Generate new result from new agents: $\{R_{(0,t)}, R_{(1,t)}, ..., R_{(N-1,t)}\} \leftarrow \{A_{(0,t)}, A_{(1,t)}, ..., A_{(N-1,t)}\}$
    
    Integrate their results as a natural selection processing: $R_{t} \leftarrow \mathbf{Evo}_{\texttt{Update}}(\{R_{(0,t)}, R_{(1,t)}, ..., R_{(N-1,t)}\}, R_{t-1})$\;
    
}

\Return $R_T \leftarrow R_t$;
\end{algorithm*}

\subsection{Preliminary}\label{sec:ea}
Evolutionary algorithm (EA)~\citep{bartz2014evolutionary,eiben2015evolutionary}, is a general algorithm to simulate the biological behaviors in evolution, including reproduction, mutation, recombination, and selection. 
By introducing genetic algorithm~\citep{sampson1976adaptation,holland1992adaptation,mitchell1998introduction,schmitt2001theory,mirjalili2020genetic} of the ``survival of the fittest'' mechanism, it can also be considered as an optimization method to improve individuals. 
Therefore, EAs also belong to the non-parametric learning method, which can be applied to any framework. 
All we need to do is define which parts should be evolved and the corresponding evolutionary operators. 
We also note some recent works~\cite{Qingyan2023Evoprompt, Angelica2023Evoprompting} indicate the potential of EAs that can be applied to optimize discrete prompts. 
So, in this paper, we explore how to formulate the agent generation problem as an evolutionary task.

\subsection{\method}\label{sec:frame} 
By assigning various settings to specific skills (\eg, role-playing, planning, tools and so on), agents could exhibit diverse task-solving capabilities. Therefore, our objective is to produce a population of agents with distinct skills, to establish effective multi-agent systems. To fulfill this point, we treat each specialized agent as an unique individual and denote each skill as the part to be evolved, akin to humans. So, we consider the procedure of agent generation to be evolutionary processing. Specifically, 
existing frameworks usually describe agent skills as the language. 
Thus, we can employ LLM to simulate evolutionary operators to update the system settings of agents and create new agents.
As shown in Figure~\ref{figue:evoagent}, we formulate the procedure of \method as a four-stage pipeline:

\paragraph{STEP 1: Initialization} 
To conduct EAs, we first need to confirm our initial agents. 
Here, we enable \method to start from a pre-defined agent framework (\eg, MetaGPT~\cite{hong2024metagpt} and AutoGen~\cite{wu2023autogen}), which serves as the initial (parent) agents.
Moreover, we also define which parts of this agent should be upgraded. 
Generally, since EAs is a generic algorithm, \method is applicable to any agent frameworks and extends them as multi-agent frameworks.
We will then explore how to generate new agents in the next steps.

\paragraph{STEP 2: Crossover \& Mutation}
In the first iteration, we directly use the initial agents as the parents. And then, we design two kinds of evolutionary operators, named \textit{Crossover} and \textit{Mutation}. 
For \textit{Crossover}, we first enable the parent agents to generate results based on user requests. 
Then, based on the generated results, we ask LLMs to check which skills should be improved and then update them. 
This mechanism allows us to generate child agents in new settings without requiring any human labor. 
Moreover, we also need to guarantee the diversity between the child agents and parents. 
To this end, we design a \textit{Mutation} operation that requires LLMs to compare the child agents and parent agents and then modify the child agents to make them distinct from their parents while maintaining their task-solving capability. 
Based on these evolutionary operators, we can generate effective and diverse agents during one iteration. 
Besides, as we also need to conduct multiple iterations, we will append all agents generated in the previous generation into the next iteration. 

\paragraph{STEP 3: Selection}
Based on the above steps, we can obtain multiple candidate agents with diverse settings. 
To guarantee the quality of agents, we also introduce a selection mechanism like EAs. 
Here, we conduct a \textit{quality-check module} with an LLM to detect whether the generated agents can satisfy it has inherited the characteristics and maintained differences from parent agents. 
We will select N child agents as the evolved agents in each iteration.

\paragraph{STEP 4: Results Update}
Based on the above steps, we obtain many new agents that evolved from parent agents, but with diverse settings. 
To improve task-solving capabilities, we ask each child agent to generate candidate results and then use LLMs to integrate these candidates with the result from the previous iteration into a new result, akin to a natural selection processing stage.
Moreover, we can automatically generate more agents by repeating the operations from step 2 to step 4 until the number of agents has fulfilled our targets.

By introducing EA, \method enables us to automatically extend the existing agent framework to a multi-agent system without any extra human designs. 
The mechanism also makes \method can be applied to any agent framework without any prerequisites.
We also present the details of \method in Algorithm~\ref{alg:evoagent}.

\section{Experiment}
\label{sec:exevalu}

In this section, we adopt \method to multiple applications to illustrate that \method can help LLM-based agents better accomplish tasks with multi-agent generation.\footnote{The data examples of \method on these tasks are provided in Appendix~\ref{sec:method_example}.}
We also demonstrate that \method can be applicable in supporting currently widely used multi-agent frameworks, such as MetaGPT, AutoGen, and Camel in Appendix~\ref{sec:appendix_adaption}.

\subsection{NLP and Multi-Modal Tasks}\label{sec:nlp_task}
\begin{table}[t]
\small
    \caption{
    Results of LLMs with different methods on Logic Grid Puzzle (\textbf{Logic}), Trivia Creative Writing (\textbf{Writing}) and Codenames Collaborative (\textbf{Code}). 
    The best results are \textbf{bolded}, and the second best ones are \underline{underlined}.
    }
  \centering
    \begin{tabular}{llrrr}
    \toprule
    \textbf{Model} & \textbf{Method} & \textbf{Logic} & \textbf{Writing} & \textbf{Code}\\
    \midrule
    \multicolumn{1}{l}{\multirow{5}[0]{*}{LLama2-13B}} & Direct & 4.00  & 28.00  & 0.00\\
          & CoT   & 26.00  & \underline{46.00}  & \underline{18.00}  \\
          & SPP   & 0.00  & 4.00  & 1.00  \\
          & Self-Refine$_3$ & \underline{33.50}  & 31.20  & 12.37  \\
          & AgentVerse & 10.00 & 12.00 & 15.36\\
          & AutoAgents & 16.00 & 18.00 & 13.50\\
          & \textsc{EvoAgent}$_{(1,3)}$ & \textbf{35.50} & \textbf{49.60} & \textbf{27.83} \\
    \midrule
    \multicolumn{1}{l}{\multirow{5}[0]{*}{GPT-3.5}} & Direct & 48.00  & 56.20  & \underline{76.29}  \\
          & CoT   & 47.50  & 51.00  & 71.13 \\
          & SPP   & {56.00}  & 54.40  & 61.86  \\
          & Self-Refine$_3$ & 47.50  & \underline{59.19}  & 46.39  \\
          & AgentVerse & 66.50 & 56.20 & 50.48\\
          & AutoAgents & \underline{68.00} & 55.35 & 52.16\\
          & \textsc{EvoAgent}$_{(1,3)}$ & \textbf{71.50} & \textbf{60.80} & \textbf{79.38} \\
    \midrule
    \multicolumn{1}{l}{\multirow{7}[0]{*}{GPT-4}} & Direct & 60.50  & 75.40  & 79.38  \\
          & CoT   & 65.50  & 74.00  & 80.41  \\
          
          & SPP   & 64.50  & 79.20  & 78.35  \\
          & Self-Refine$_3$ & 64.50  & 74.60  & 79.38  \\
          & AgentVerse & 66.50 & 78.00 & 80.41\\
          & AutoAgents & \underline{69.00} & \underline{82.00} & \underline{83.56}\\
          & \textsc{EvoAgent}$_{(1,3)}$ & \textbf{77.00} & \textbf{84.40} & \textbf{84.53}\\
    \bottomrule
    \end{tabular}%

  \label{tab:ssp}%
\end{table}%

\paragraph{Benchmarks}
To align previous experiences, \eg, Self-Refine~\cite{madaan2023selfrefine} and Solo Performance Prompting~\cite{wang2023unleashing}, we select three NLP knowledge-intensive and reasoning-intensive tasks from \cite{wang2023unleashing} and one multi-modal task:
\begin{itemize}[leftmargin=*]
    \item \textbf{Logic Grid Puzzle} is a reasoning task with 200 puzzles featuring 2 to 5 unique occupants in different houses. The aim is to identify house numbers for one occupant with provided clues.
    \item \textbf{Trivia Creative Writing} is a knowledge-intensive task consisting of 100 instances. This task requires a model to write a coherent story while incorporating answers to N trivia questions.
    \item \textbf{Codenames Collaborative} is a reasoning-intensive task with 50 instances. It involves a model identifying target words based on a given hint and a complete list of words.
    \item \textbf{MMMU}~\cite{yue2023mmmu} is a comprehensive and general benchmark for multi-discipline multi-modal understanding and reasoning. MMMU has three levels of difficulty: easy, medium, and hard. We evaluate \method against baselines using the multiple-choice questions in the validation set of MMMU, which includes 847 questions spanning 30 different domains.
\end{itemize}

\paragraph{Baselines}
For NLP tasks, we select LLama2-13B-Chat~\cite{Hugo2023LLaMa2}, GPT-3.5~\cite{openai2022chatgpt} and GPT-4~\cite{OpenAI2023GPT4} as our backbone networks.
We compare \method with 0-shot learning (Direct), Chain-of-thought (CoT) prompting~\cite{wei2022chain} and Self-Refine~\cite{madaan2023selfrefine} and Solo Performance Prompting (SPP)~\cite{wang2023unleashing}.
For Self-Refine, we follow \cite{madaan2023selfrefine} to design feedback and refine prompts with three iterations.
SPP is a multi-agent collaboration prompting strategy that asks a single LLM to identify and discuss with multiple personas with few-shot learning.
For SPP, we follow the original setting~\cite{wang2023unleashing}.
We also compare \method with some pre-defined multi-agent agent generation frameworks, \ie, AgentVerse~\cite{chen2024agentverse} and AutoAgent~\cite{chen2023autoagents}.
For MMMU, we select GPT-4V~\cite{yang2023dawn} and Gemini-Pro as the backbone and compare \method with CoT prompting, Self-Refine (SR), and SPP.\footnote{The detailed model parameters 
model versions, baseline introduction and full prompts for these methods can be found in Appendix~\ref{sec:exp_appendix}.}

\paragraph{Evaluation Metrics}
We adhere to the evaluation metrics specified in the original setting. 
Specifically, for Logic Grid Puzzle and MMMU tasks, we report the accuracy of all questions.
For Trivia Creative Writing, we measure the ratio of correctly mentioned answers in the trivia questions.
For Codenames Collaborative, we calculate the overlapping ratio between the predicted words from the Guesser and the target words as the metric.

\begin{figure}[t]
    \centering
\includegraphics[width=\linewidth]{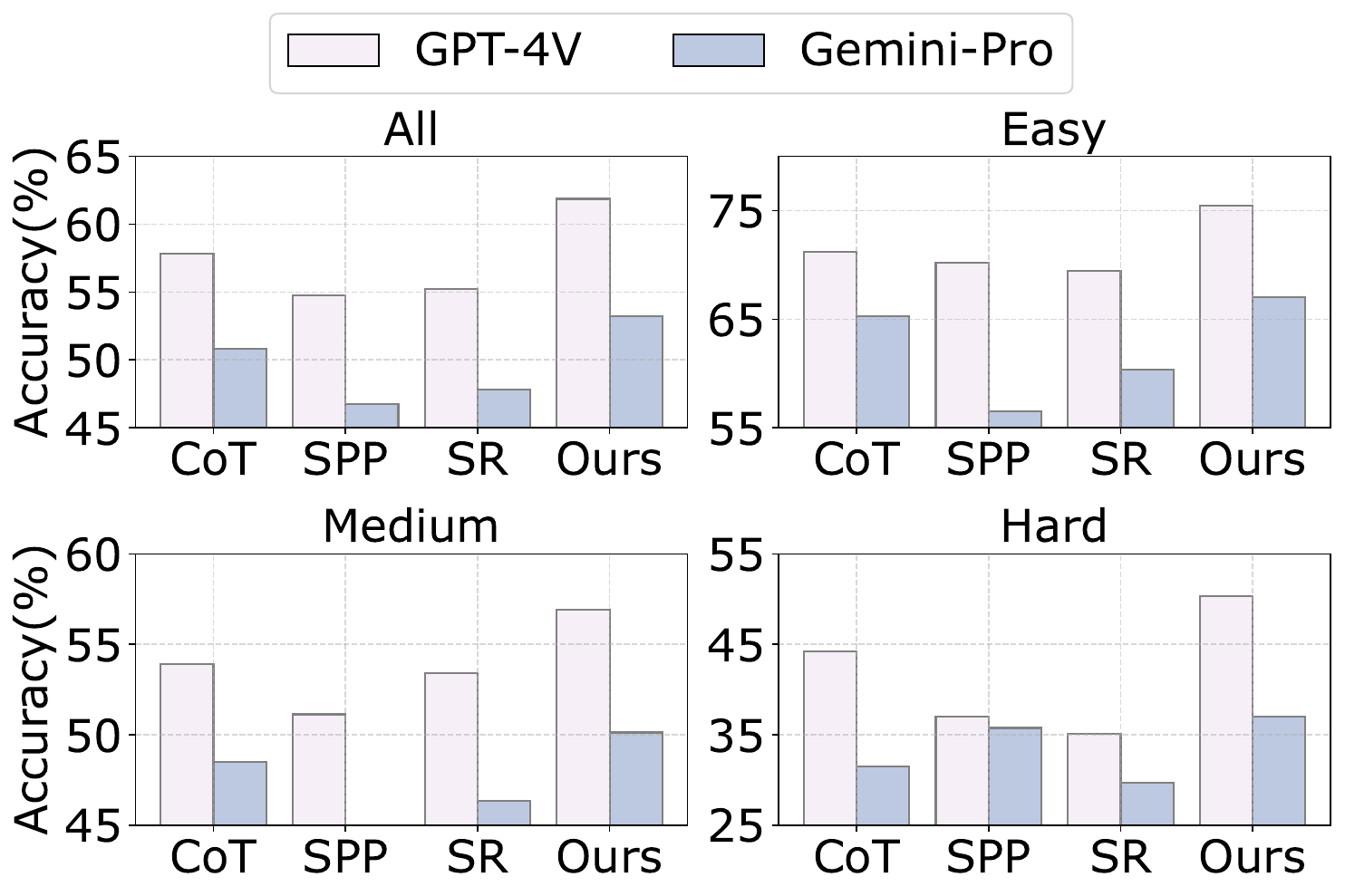}
    \caption{Overall results of GPT-4V and Gemini-Pro with different methods on the MMMU validation set. We also compare the performance of GPT-4V and Gemini-Pro across three difficulty levels.
    }
    \label{fig:mmmu}
\end{figure}

\paragraph{Result \& Analysis}
In our experiments, we adopt the agent settings of \cite{wang2023unleashing} (for NLP tasks) and \cite{yue2023mmmu} (for MMMU) as the initial agent.
For our method, we denote it as \textsc{EvoAgent}$_{(N,T)}$, where N is the population size generated in each iteration, and T is the number of iterations. 
Here, to align with Self-Refine, we set N as 1 and T as 3, which means we conduct three iterations, each of which generates a new specialized agent.
Thus, for each data sample, \method extends 3 different specialized agents to collaborate with the initial agent.
Our results are reported in Table~\ref{tab:ssp}, and we can observe: 
\begin{enumerate}[leftmargin=*]
    \item By utilizing multiple generated agents, \method can greatly improve LLM performances in both NLP knowledge and reasoning tasks. 
    Moreover, \method outperforms both AgentVerse and AutoAgent, highlighting the effectiveness and generality of \method.
    \item When using weaker LLMs, SPP usually produces poor performances, consistent with the findings in \cite{wang2023unleashing}.
    This suggests the limited effectiveness of SPP in smaller and less capable models.
    However, \method can provide consistent improvements among each LLM, proving its strong generalization by using diverse generated agents. 
\end{enumerate}
In addition, Figure~\ref{fig:mmmu} shows that Self-Refine (SR) and SPP degrade performance compared to CoT prompting in MMMU task.
However, \method can generate multiple domain-specific agents and thus improve multi-modal models in addressing scientific questions across various difficulty levels.
More analysis about the time efficiency of \method is shown in Appendix~\ref{appendix:efficiency}.

\subsection{Interactive Scientific Solving Simulation}\label{sec:scienceworld}

\paragraph{Benchmark}
Compared with traditional NLP or multi-modal tasks, autonomous agents usually need to perform problem-solving abilities akin to humans in interactive and open-world environments. 
We choose ScienceWorld~\cite{wang-etal-2022-scienceworld}, a complex interactive environment requiring skills in long-term memory, sub-task decomposition, and scientific and commonsense knowledge. 
We evaluate 30 scientific tasks in ScienceWorld to demonstrate the capability of \method in solving tasks in more challenging open-world environments.

\begin{table}[t]
\small
\caption{Average Scores of different methods on ScienceWorld. We also report performance on three difficult-level groups based on the average length of the oracle agent’s trajectories~\citep{lin2023swiftsage}. 
    }
    \vspace{0.15cm}
  \centering
    \begin{tabular}{lrrrr}
    \toprule
    \textbf{Model} & \textbf{Overall} & \textbf{Long} & \textbf{Medium} & \textbf{Short} \\
    \midrule
    GPT-3.5 & 17.12 & 6.28  & \textbf{19.91} & 27.90 \\
    \ \ w/ \textsc{EvoAgent}$_{(1,1)}$ & \textbf{19.02} & \textbf{7.25} & 18.87 & \textbf{33.26} \\
    \midrule
    GPT-4 & 27.97 & 10.58 & 36.00    & 42.41 \\
    \ \ w/ \textsc{EvoAgent}$_{(1,1)}$ & \textbf{30.42} & \textbf{11.38} & \textbf{36.17} & \textbf{48.67} \\
    \bottomrule
    \end{tabular}%
    
  \label{tab:scienceworld}%
\end{table}

\begin{table*}[t]
\small
\caption{Main results of different LLMs and planning strategies on the TravelPlanner validation set. 
    \textsc{EvoAgent}$_{(N,T)}$ indicates that the population size per iteration is N and the number of iterations is T.
    The best results are \textbf{bolded}, and the second best ones are \uline{underlined}.}
  \centering
    \begin{tabular}{llcccccc}
    \toprule
    \multicolumn{1}{l}{\multirow{2}[0]{*}{\textbf{Model}}} & \multirow{2}[0]{*}{\textbf{Method}} & \textbf{Delivery}  & \multicolumn{2}{c}{\textbf{Commonsense}} & \multicolumn{2}{c}{\textbf{Hard Constraint}} & \multirow{2}[0]{*}{\textbf{Final}} \\
    \cmidrule(lr){4-5}
    \cmidrule(lr){6-7}
          &       &    \textbf{Rate}  & \textbf{Micro} & \textbf{Macro} & \textbf{Micro} & \textbf{Macro} &  \\
    \midrule
    \multicolumn{1}{l}{\multirow{4}[0]{*}{Mistral-7B}} & Direct & 100.0  & \textbf{64.7} & \textbf{2.2} & \underline{3.1}   & 0.0   & 0.0  \\
          & CoT   & 100.0  & \underline{60.5}  & \underline{1.1}   & 1.0   & 0.0   & 0.0  \\
          & SPP   & 100.0  & 55.1  & 0.0   & 0.7   & \textbf{0.6} & 0.0  \\
          & Self-Refine$_3$ & 100.0  & 58.3  & 0.0   & 0.7   & 0.0   & 0.0  \\
          
          & \textsc{EvoAgent}$_{(1,3)}$ & 100.0  & 60.1  & \textbf{2.2} & \textbf{4.5} & \textbf{0.6} & 0.0  \\
    \midrule
    \multicolumn{1}{l}{\multirow{7}[0]{*}{GPT-3.5}} & Direct & 100.0  & 57.3  & \underline{3.9}   & 11.0  & 3.3   & 0.0  \\
          & CoT   & 100.0  & \underline{61.0}  & 2.8   & 10.0  & 3.3   & 0.0  \\
          & ReAct & 82.2  & 42.3  & 0.6   & \textbf{11.9}  & \textbf{4.6}   & 0.0  \\
          & SPP   & 99.4  & 54.6  & 1.7   & 3.8   & 1.1   & 0.0  \\
          & Self-Refine$_3$ & 100.0  & 56.0  & 1.7   & 3.1   & 1.1   & 0.0  \\
          
          & \textsc{EvoAgent}$_{(1,3)}$ & 100.0  & \textbf{64.2} & \textbf{7.8} & \underline{11.0}  & \underline{4.4}   & \textbf{1.1} \\
    \midrule
    \multicolumn{1}{l}{\multirow{6}[0]{*}{Gemini-Pro}} & Direct & 90.0  & 61.7  & \underline{7.8}   & \underline{16.4}  & \textbf{7.8}   & \underline{0.6}  \\
          & CoT   & 90.0  & 61.4  & 7.2   & 10.0  & 6.1   & \textbf{1.7}  \\
          & SPP   & 100.0  & \underline{67.6}  & \underline{7.8}   & 10.2  & 3.9   & 1.1  \\
          & Self-Refine$_3$ & 95.6  & 65.8  & 6.1   & 15.0  & 4.4   & \underline{0.6}  \\
          
          & \textsc{EvoAgent}$_{(1,3)}$ & 100.0  & \textbf{73.5}  & \textbf{12.8} & \textbf{16.9}  & \underline{7.2}   & \textbf{1.7}  \\
    \midrule
    \multicolumn{1}{l}{\multirow{4}[0]{*}{GPT-4}} & Direct & 100.0  & \underline{79.4}  & \underline{15.8}  & \underline{27.5}  & \underline{16.1}  & \underline{2.2}  \\
          & CoT   & 100.0  & 76.7  & 11.7  & 22.4  & 12.8  & \underline{2.2}  \\
          & SPP   & 96.7  & 70.6  & 5.6   & 11.4  & 7.8   & 0.6  \\
          & Self-Refine$_3$ & 98.9  & 75.3  & 7.2   & 12.4  & 7.2   & 1.1  \\
          
          & \textsc{EvoAgent}$_{(1,3)}$ & 100.0 & \textbf{81.5} & \textbf{21.1} & \textbf{31.4} & \textbf{18.9} & \textbf{7.2} \\
    \bottomrule
    \end{tabular}%
    
  \label{tab:travelplanner}%
\end{table*}%

\paragraph{Baseline and Evaluation Metrics}
Following \cite{lin2023swiftsage}, we require LLMs to perform an action at each step by using in-context learning~\footnote{The introduction of the settings of LLMs are shown in Appendix~\ref{sec:baseline_sw}.}.
For evaluation, each task in ScienceWorld includes some sub-tasks, and we report the results by calculating the completed sub-tasks for the whole task.

\paragraph{Result \& Analysis}

For \method, we adopt the agent framework with original settings in \cite{lin2023swiftsage} as the initial agent. 
Since each step in ScienceWorld requires using EA, we set the population size N as 1 and the iterations T as 1 for efficiency, denoted as \textsc{EvoAgent}$_{(1,1)}$.
Thus, for each task in ScienceWorld, \method can extend M different specialized agents to collaborate with the initial agent, where M is the number of steps in this task.
Results in Table~\ref{tab:scienceworld} show that: 
\begin{enumerate}[leftmargin=*]
    \item \method can also extend interactive agents to multi-agent systems in solving complete scientific tasks in dynamic, open-world environments and consistently improve the performance of LLMs.
    \item Our method exhibits the most substantial improvement in short-trajectory tasks, with less significant gains in medium and long-trajectory tasks.
    We argue that the capability of multi-agent systems will also be affected by a longer context.
    Future work can investigate the effect of long context on multi-agent systems.
\end{enumerate}
Generally, these results also demonstrate the generalization of \method, which can also be used for solving interactive tasks in an open-world environment.

\subsection{Real-World Scenarios}\label{sec:travelplanner}

\paragraph{Benchmark}
Planning in complex and realistic environments is also a crucial skill for building autonomous agents. 
Thus, we also select TravelPlanner~\cite{xie2024travelplanner}, a benchmark designed to evaluate language agents in real-world complex planning with multiple constraints.

\paragraph{Baseline and Evaluation Metrics}
Following \cite{xie2024travelplanner}, we select Mistral-7B~\cite{jiang2023mistral}, GPT-3.5, Gemini-Pro~\cite{geminiteam2023gemini} and GPT-4 as our backbone models.
We compare \method with 0-shot learning (Direct), CoT prompting, ReAcT~\cite{yao2023react}, SPP, and Self-Refine within each backbone model.
Furthermore, we also attempt the ReAcT method~\cite {yao2023react} for GPT-3.5, which introduces a virtual `think' action to generate sub-tasks during the action planning process.
For evaluation, we adhere to the original metrics from TravelPlanner, reporting the delivery rate, commonsense constraint pass rate, hard constraint pass rate, and final pass rate for all methods~\footnote{Detailed introduction of experiment settings is provided in Appendix~\ref{sec:tp_appendix}.}.

\paragraph{Result \& Analysis}
For \method, we adopt the original settings in TravelPlanner as the initial agent.
In our main experiment, to fairly compare with Self-Refine, we set the population size N as 1 and the iterations T as 3 for efficiency, denoted as \textsc{EvoAgent}$_{(1,3)}$.
Thus, for each user query in TravelPlanner, \method can extend 3 different specialized agents to collaborate with the initial agent.
Results in Table~\ref{tab:travelplanner} show that:
\begin{enumerate}[leftmargin=*]
    \item Although existing paradigms (\eg, Self-Refine and SPP) have demonstrated decent results in some conventional NLP tasks, they still lack capability in handling complex planning tasks. These results also demonstrate that only using human-design prompting strategies is insufficient to handle complex planning tasks.
    \item \method can automatically generate multiple agents, such as those focused on culinary experiences and transportation, and forming a multi-agent collaboration paradigm. 
    Therefore, the generated travel plans are more aligned with user preferences and commonsense rules.
    \item By using \method to automatically generate multiple agents and forming a multi-agent collaboration paradigm, we can develop higher-quality plans that better meet user preferences. That also indicates the significance of multi-agent systems for complex planning tasks.
\end{enumerate}

\subsection{Ablation Studies}
To better understand the value of \method, we conduct detailed analyses on TravelPlanner, focusing on the impact of population size and iteration number and the effectiveness of the quality-check module in the selection stage.

\begin{table}[t]
\small
\caption{Average commonsense constraint pass rate (\textbf{Com.}) and hard constraint pass rate (\textbf{Hard}) of ablated variants on TravelPlanner.
  }
  \vspace{0.1cm}
  \centering
    \begin{tabular}{lcccc}
    \toprule
    \multicolumn{1}{l}{\multirow{2}[0]{*}{\textbf{Method}}} & \multicolumn{2}{c}{\textbf{w/o QC}} & \multicolumn{2}{c}{\textbf{w/ QC}} \\
    \cmidrule(lr){2-3}
    \cmidrule(lr){4-5}
           & \multicolumn{1}{c}{\textbf{Com.}} & \multicolumn{1}{c}{\textbf{Hard}}  & \multicolumn{1}{c}{\textbf{Com.}} & \multicolumn{1}{c}{\textbf{Hard}} \\
    \midrule
     Direct &  - & -  & 59.5  & 13.7  \\
     Suggest$_3$  & - & -  & 61.7  & 8.4  \\
     Overgen$_3$  & - & -  & 61.4  & 10.7  \\
     PromptRefine$_3$  & - & -  & 63.0  & 13.8  \\
     \midrule
     \rowcolor[gray]{0.95}\multicolumn{5}{c}{\textit{Different Population Size}} \\
    \textsc{EvoAgent}$_{(1,3)}$ & \textbf{68.9} & 14.0 & \textbf{68.9} & 14.0  \\
    \textsc{EvoAgent}$_{(1,5)}$ & 67.5 & \textbf{16.9} & 67.5 & \textbf{16.9}  \\
    \textsc{EvoAgent}$_{(2,3)}$   & 62.8  & 12.7  
    & 67.0  & 15.2  \\
    \textsc{EvoAgent}$_{(3,3)}$   & 62.7  & 13.7  
    & 66.8  & 15.8 \\
    \midrule
    \rowcolor[gray]{0.95}\multicolumn{5}{c}{\textit{Different Selection Stategies}} \\
    Random 
    & 62.9  & 12.7 
    & \textbf{67.1}  & 15.0  \\
    PK 
    & \textbf{63.5}  & \textbf{13.6} 
    & 66.4  & 14.5  \\
    All-in 
    & 61.9  & 13.2  
    & \textbf{67.1}  & \textbf{17.0} \\
    \bottomrule
    \end{tabular}%
  
  \label{tab:ablation_small}%
\end{table}


\paragraph{Experiment Settings}
We evaluate the performance of different LLMs at varying population sizes N and iteration number T.
For each query, \method can generate N$\times$T different specialized agents to collaborate with the initial agent.
We employ an LLM that shares the same backbone as the initial agent for updates.
To select results from candidates for this LLM to update, we adopt three different selection strategies:
1) \textit{Random}: one result is selected randomly from the pool of candidates;
2) \textit{PK}: we ask an agent with the same backbone as the initial agent to identify the optimal results from the pool of candidates; 
3) \textit{All-in}: Rather than selecting a single result, we update using all candidates.

Moreover, we also attempt Suggest$_{3}$, Overgen$_{3}$ and PromptRefine$_{3}$ as variants to prove the effectiveness of our method.
For Suggest$_{3}$, instead of generating new results, we ask new generated agents to only give suggestions for initial agents to revise their results. 
For Overgen$_{3}$, we first ask initial agents to generate 3 different results at one time, and then these agents can output the final results based on these multiple candidates.
For PromptRefine$_{3}$, instead of generating agents, we ask the initial agent to refine its prompts three times to better answer the query.
\footnote{The full prompts of different ablation settings are shown in Appendix~\ref{sec:prompt_design}.}

\paragraph{Result \& Analysis}
To obtain stable findings, we first obtain results from GPT-3.5 and Gemini-Pro across different population sizes and selection strategies. 
We then average their results over various metrics to clearly compare the strengths and weaknesses of these variants.

The results are shown in Table~\ref{tab:ablation_small}.\footnote{The complete results with further analysis are shown in Appendix~\ref{sec:ab_appendix}} 
We find that \method significantly outperforms the Overgen, demonstrating the effectiveness of generating specialized agents to assist with complex planning.
Although obtaining suggestions from new generated agents can improve the performance on commonsense constraints, these methods greatly harm the agents to meet the user preference.
Modifying the prompt can improve the performance of agents, yet it remains less effective than \method.

When the population size exceeds one (N$>1$), agents may generate similar agents.
Thus, lacking a quality-check module leads to reduced travel plan quality.
Furthermore, when population size and iteration number increase, the model aligns travel plans more closely with user preferences but diminishing adherence to commonsense rules, consistent with the findings in Table~\ref{tab:travelplanner}.
We hypothesize that this variability stems from the initial agent's bias in adjusting its outputs based on the results generated by new agents, notably prioritizing user preferences over commonsense rules. 
Future work can explore and alleviate this bias.

Remarkably, the PK strategy initially yields superior results without the quality-check module, but this trend reverses once quality checks are implemented.
We speculate that, without the quality-check module, PK partially fulfills this role, aiding in selecting better candidates. 
However, with the quality-check module, PK introduces bias by favoring specific fields of expertise while neglecting others, resulting in a less effective than random strategy.
Meanwhile, the All-in strategy performs optimally when a quality-check module is included.
Future research can leverage long-context LLMs to expand more agents with \method to better solve complex real-world tasks.

\section{Conclusion}
\label{sec:conclusion}
In this paper, we propose \method, an automatic multi-agent generation system by leveraging evolutionary algorithms. \method is suitable to any existing agent framework and extends it to multi-agent systems with diverse and effective agents by using a series of evolutionary operations, including mutation, crossover, and selection. 
Experiments on multiple tasks show that \method can significantly improve the capabilities of LLM-based agents in solving complex tasks.

\section*{Limitations}
\label{sec:limitation}

First, \method requires the model to generate multiple specialized agents, which brings more token cost than a single agent.
Besides, except for AgentVerse and AutoAgents in Table~\ref{tab:ssp}, we do not conduct extensive comparisons with existing multi-agent systems frameworks.
This is because these frameworks require additional design efforts, and designing a suitable framework for each benchmark used in our experiment is beyond the scope of this paper.
Finally, we do not manually evaluate the quality of the generated agents, but the results show that \method significantly improves the performance of LLM-based agent on complex tasks.

\section*{Ethics Statement}
\label{sec:Ethics}

We acknowledge that all authors are informed about and adhere to the ACL ARR Code of Ethics and the Code of Conduct.

\paragraph{Risks}
The benchmarks in our experiment are sourced from publicly available sources. However, we cannot guarantee that they are devoid of socially harmful or toxic language. 
We use ChatGPT~\cite{openai2022chatgpt} to correct grammatical errors in this paper.

\section*{Acknowledgement}
We thank the anonymous reviewers for their valuable comments. This work is partially supported by the Chinese NSF Major Research Plan (No.92270121).

\bibliography{anthology,custom}

\clearpage
\appendix
\begin{appendix}
\label{sec:appendix}

\section{Experiment Settings}\label{sec:exp_appendix}

\subsection{Prompt for Baselines and \method}\label{sec:prompt_design}
Listing~\ref{lst:self-refine} and \ref{lst:ssp} shows the full prompt for 0-shot learning (Direct), Chain-of-thought (CoT) prompting~\citep{wei2022chain} and Self-Refine~\citep{madaan2023selfrefine} and Solo Performance Prompting, \ie, SPP~\citep{wang2023unleashing}.
Listing~\ref{lst:instruction} and \ref{lst:ablation} show the prompt of \method and different ablation settings.

\subsection{Model Selection}\label{sec:model_selection}
For OpenAI models, we use \texttt{gpt-35-turbo} and \texttt{gpt-4-32k} with the version of \texttt{2024-02-15-preview} in Azure.\footnote{\url{https://azure.microsoft.com/en-us/products/ai-services/openai-service}}
For Gemini-pro, we use Google Gemini-Pro APIs to obtain results.
We set the temperature to 0 for all models.

\subsection{Human-designed Agent Framework}\label{Appendix:human-designed_agent}

AgentVerse~\citep{chen2024agentverse} and AutoAgent~\citep{chen2023autoagents} are frameworks designed to generate an unlimited number of agents for collaborative tasks automatically. Despite this automation, they still rely on human-designed interventions. 
AutoAgents requires agent settings to satisfy a ``Planner - Agent Observer - Plan Observer'' framework, while AgentVerse formulates a pipeline of ``Expert Recruitment - Collaborative Decision Making - Action Execution - Evaluation'' to build agents.
We argue that these human-designed architectures limit their scalability and functionality.
We follow their experimental settings and compared them with our method.

\subsection{Experimental Details of ScienceWorld}\label{sec:baseline_sw}
Following \cite{lin2023swiftsage}, we adopt the REACT~\citep{yao2023react} method for each LLM, which introduces a virtual 'think' action. This action allows LLMs to generate subgoals during the action planning process.

\subsection{Evaluation Details of TravelPlanner}\label{sec:tp_appendix}

Grounding to travel planning, a real-world use-case that inherently involves various constraints like user preferences and commonsense rules, TravelPlanner evaluates whether agents can formulate flexible travel plans using gathered information to meet these constraints.
We test \method and all baselines on the TravelPlanner validation set, which consists of 180 user queries with the collected information.
To evaluate the travel plans generated by agents, TravelPlanner adopts the following evaluation metrics: 
\begin{itemize}[leftmargin=*]
    \item Delivery Rate: Assesses if agents can complete a plan within a limited number of steps (30 in our experimental setting). Failures are due to dead loops, numerous failed attempts, or exceeding the step limit.
    \item Commonsense Constraint Pass Rate: Evaluates if an agent can incorporate commonsense into their plan.
    \item Hard Constraint Pass Rate: Measures if a plan meets all explicit hard constraints in the query, testing the agent's ability to adapt to diverse user preferences.
    \item Final Pass Rate: Indicates the proportion of viable plans that meet all criteria, reflecting the agent's proficiency in creating practical plans.
\end{itemize}

Furthermore, TravelPlanner uses micro and macro strategies to assess the Commonsense and Hard Constraint Pass Rates. 
The micro strategy calculates the ratio of met constraints to the total. 
The macro strategy measures the proportion of plans that meet all commonsense or hard constraints. 
Together, these strategies assess an agent's ability to satisfy individual constraints and all constraints comprehensively.

\section{More Analysis of \method}
\begin{table*}[t]
    \centering
    \small
    \caption{Time cost of different methods for Llama-3.1-70B-Instruct.}
    \begin{tabular}{lcccccc}
    \toprule
        \textbf{Method} & \textbf{Logic} & \textbf{Logic Time (s)} & \textbf{Writing} & \textbf{Writing Time} (s) & \textbf{Code} & \textbf{Code Time (s)} \\ 
    \midrule
        Direct & 50.00 & 318.00 & 56.40 & 441.00 & 64.32 & 168.00 \\ 
        CoT & 53.00 & 443.40 & 63.20 & 628.80 & 62.74 & 225.00 \\ 
        SPP & 55.50 & 648.00 & 58.00 & 822.00 & 59.26 & 445.20 \\ 
        Self-Refine & 53.00 & 942.00 & 60.80 & 1098.00 & 59.26 & 628.20 \\ 
        AgentVerse & 60.00 & 1056.00 & 70.40 & 1518.00 & 68.90 & 824.40 \\ 
        AutoAgents & 61.50 & 988.80 & 69.60 & 1303.80 & 68.35 & 770.40 \\ 
    \midrule
        EvoAgent & \textbf{67.00} & 1060.80 & \textbf{74.50} & 1209.60 & \textbf{74.62} & 859.20 \\ 
    \bottomrule
    \end{tabular}
    \label{tab:efficiency}
\end{table*}
\subsection{Time Efficiency of \method}\label{appendix:efficiency}
We conduct experiments on Logic Grid Puzzle (Logic), Trivia Creative Writing (Writing), and Codenames Collaborative (Code) by using Llama-3.1-70B-Instruct~\cite{dubey2024llama} with SGLang~\footnote{\url{https://github.com/sgl-project/sglang}}.
SGLang is a fast-serving framework that allows us to track the time and computational resources required for model improvements.
Specifically, we deploy Llama-3.1-70B-Instruct on an 8-GPU setup using RTX 3090 (bf16).
We compare \method with a single-agent approach and other multi-agent systems. 
The results are presented in Table~\ref{tab:efficiency}.
Our findings show that, compared to a single-agent system, \method requires more time but delivers better performance. 
When compared to multi-agent systems such as AgentVerse and AutoAgents, 
\method achieves similar time costs while producing the best performance. These results highlight the effectiveness and versatility of \method.

\subsection{More Analysis of Ablation Studies}\label{sec:ab_appendix}
\begin{table*}[t]
  \centering
  \caption{Comparison of different popularity selection strategies for LLMs on TravelPlanner. The best results are \textbf{bolded}, and the second best ones are \uline{underlined}.}
  \small
    \begin{tabular}{lllcccccc}
    \toprule
    \multirow{2}[0]{*}{\textbf{Model}} & \multirow{2}[0]{*}{\textbf{Strategy}} & \multirow{2}[0]{*}{\textbf{Method}} & \multicolumn{3}{c}{\textbf{w/o Quality Check}} & \multicolumn{3}{c}{\textbf{w/ Quality Check}} \\
    \cmidrule(lr){4-6}
    \cmidrule(lr){7-9}
         &       &       & \textbf{Delivery} & \multicolumn{1}{c}{\textbf{Com.}} & \multicolumn{1}{c}{\textbf{Hard}} & \textbf{Delivery} & \multicolumn{1}{c}{\textbf{Com.}} & \multicolumn{1}{c}{\textbf{Hard}} \\
    \midrule
    \multirow{11}[0]{*}{GPT-3.5} &  & Direct &  - & -  & - & 100.0  & 57.3  & 11.0  \\
        &  &  Suggest$_3$ & -     & -     & -     & 100.0  & 57.5  & 5.7  \\
          &  &  Overgen$_3$ &  - & -  & - & 98.3  & 56.3  & 9.0  \\
          &  &  PromptRefine$_3$ &  -     & -     & -     & 100.0  & 61.2  & 11.0  \\
          \cmidrule(lr){2-9}
          &  & \textsc{EvoAgent}$_{(1,3)}$ & 100.0  & \textbf{64.2} & 11.0  & 100.0  & 64.2  & 11.0  \\
          & & \textsc{EvoAgent}$_{(1,5)}$ & 100.0  & \underline{61.0}  & \textbf{12.6}  & 100.0  & {61.0}  & {12.6}  \\
          \cmidrule(lr){2-9}
          & \multirow{2}[0]{*}{Random} & \textsc{EvoAgent}$_{(2,3)}$ & 100.0  & 59.4  & 10.2  & 100.0   & 65.4  & 13.8 \\
          &       & \textsc{EvoAgent}$_{(3,3)}$ & 98.9  & 59.2  & \underline{11.4} & 100.0  & \underline{65.8}  & \underline{14.0}  \\
          \cmidrule(lr){2-9}
          & \multirow{2}[0]{*}{PK} & \textsc{EvoAgent}$_{(2,3)}$ & 99.4  & 59.4  & 7.1   & 100.0  & \textbf{66.0} & 11.7  \\
          &       & \textsc{EvoAgent}$_{(3,3)}$ & 98.9  & 58.5  & {11.2}  & 100.0  & 61.3  & 12.4  \\
          \cmidrule(lr){2-9}
          & \multirow{2}[0]{*}{All-in} & \textsc{EvoAgent}$_{(2,3)}$ & 97.2  & 59.4  & 10.0  & 100.0  & 64.2  & \textbf{15.5} \\
          &       & \textsc{EvoAgent}$_{(3,3)}$ & 93.3  & 56.0  & 8.3   & 100.0  & 65.2  & 12.6  \\
    \midrule
    \multirow{11}[0]{*}{Gemini-Pro} &  & Direct &  - & -  & - & 90.0  & 61.7  & 16.4  \\
            &  &  Suggest$_3$ & -     & -     & -     & 100.0  & 65.8  & 11.0  \\

          &  &  Overgen$_3$ &  - & -  & - & 100.0  & 66.5  & 12.4  \\
          &  &  PromptRefine$_3$ &  -     & -     & -  &   96.7  & 64.9  & 16.7  \\
          \cmidrule(lr){2-9}
          &  & \textsc{EvoAgent}$_{(1,3)}$ & 100.0  & \underline{73.5}  & 16.9 &  100.0  & \underline{73.5}  & 16.9   \\
          & & \textsc{EvoAgent}$_{(1,5)}$ & 100.0  & \textbf{74.0} & \textbf{21.2}  &  100.0  & \textbf{74.0} & \textbf{21.2} \\
          \cmidrule(lr){2-9}
          & \multirow{2}[0]{*}{Random} & \textsc{EvoAgent}$_{(2,3)}$ & 96.7  & 65.9  & 13.1  & 99.4  & 67.3  & 14.0  \\
          &       & \textsc{EvoAgent}$_{(3,3)}$ & 97.2  & 67.0  & 16.0  & 100.0  & 70.0  & {18.1}  \\
          \cmidrule(lr){2-9}
          & \multirow{2}[0]{*}{PK} & \textsc{EvoAgent}$_{(2,3)}$ & 97.2  & 67.4  & \underline{19.0} & 99.4  & 69.8  & 17.1  \\
          &       & \textsc{EvoAgent}$_{(3,3)}$ & 97.2  & {68.5} & 17.1  & 99.4  & 68.4  & 16.7  \\
          \cmidrule(lr){2-9}
          & \multirow{2}[0]{*}{All-in} & \textsc{EvoAgent}$_{(2,3)}$ & 95.0  & 65.1  & 16.7  & 99.4  & 69.0  & 19.0  \\
          &       & \textsc{EvoAgent}$_{(3,3)}$ & 95.0  & 66.9  & {17.9}  & 100.0  & {70.1} & \underline{20.7} \\
    \bottomrule
   \end{tabular}%
    
  \label{tab:ablation}%
\end{table*}%

\begin{table}[t]
    \centering
    \includegraphics[width=0.8\linewidth]{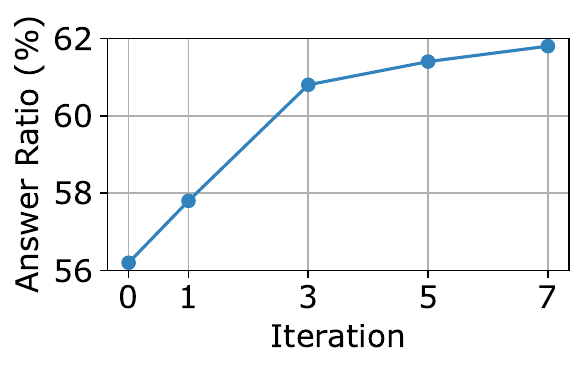}
    \caption{The performance of GPT-3.5 with \method under different iterations on Trivia Creative Writing task.}
    \label{figue:ssp_plot}
\end{table}


The complete results of ablation studies on TravelPlanner are shown in Table~\ref{tab:ablation}.
This result indicates that the absence of the quality-check module significantly lowers the delivery pass rate when the All-in strategy is applied.
To explore the reasons, we revisit the results and discover that sometimes unsuitable agents create overly lengthy travel plans that fail to meet the criteria.
For example, the model might erroneously assign a nutritionist to devise travel plans, resulting in excessively detailed meal arrangements and nutritional breakdowns.
Therefore, the input length surpasses the context window of LLMs, preventing the final result generation.

Moreover, we also conduct experiments on the Trivia Creative Writing task to investigate the impact of the number of iterations on model performance in traditional NLP tasks. 
As shown in Figure~\ref{figue:ssp_plot}, model performance improves with increasing iterations. However, the improvement plateaus when the iteration count exceeds three. We suggest that traditional NLP tasks are relatively simple, and beyond a certain iteration number, even with a quality-check module in place, the generated agents tend to be similar and thus converge.

\begin{figure*}[t]
    \centering
    \includegraphics[width=\linewidth]{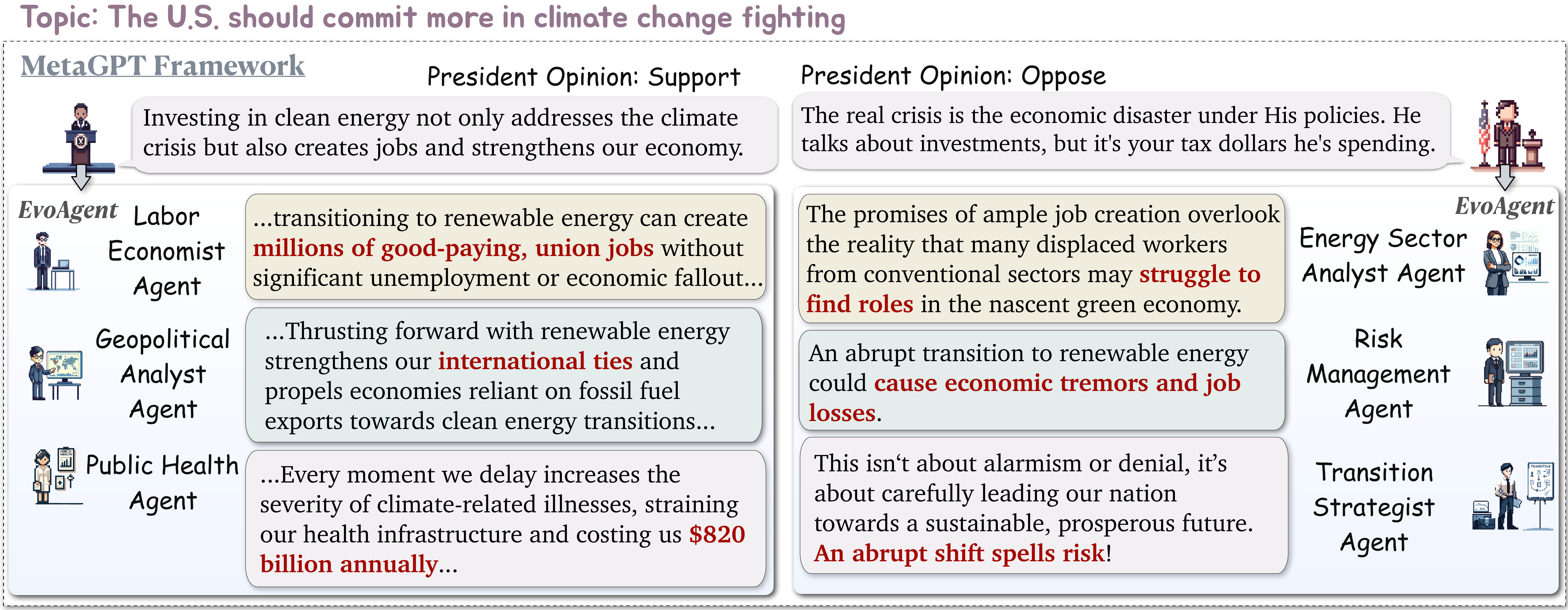}
    \caption{
    The adaption of \method on MetaGPT framework. With the EA, we can extend the original role in the debate scenario to different specialized agents to enrich the opinions.
    }
    \label{figue:metagpt}
\end{figure*}

\section{Examples of \method}\label{sec:method_example}
\subsection{\method Examples of NLP reasoning and knowledge tasks}\label{sec:example_ssp}

Listing~\ref{lst:example_logic}, \ref{lst:example_tcw} and\ref{lst:example_code} presents some multi-agent generation examples generated by GPT-4 based \method in Logic Grid Puzzle, Trivia Creative Writing and Codenames Collaborative for a better understanding.

\subsection{\method Examples of MMMU}\label{sec:example_mmmu}
Listing~\ref{lst:example_mmmux} presents some multi-agent generation examples generated by GPT-4 based \method in MMMU dataset for a better understanding.

\subsection{\method Examples of ScienceWorld}\label{sec:example_scienceworld}
Listing~\ref{lst:scienceworld} presents some multi-agent generation examples generated by GPT-4 based \method in ScienceWorld for a better understanding.

\subsection{\method Examples of TravelPlanner}\label{sec:example_travelplanner}
Listing~\ref{lst:example_travelplanner} presents some multi-agent generation examples generated by GPT-4 based \method in TravelPlanner for a better understanding.

\section{Examples of \method's Adaption to Multi-agent Collaboration Frameworks}\label{sec:appendix_adaption}

Previous experiments have demonstrated that our method can automatically extend existing agent frameworks to multi-agent systems, thus greatly improving LLM-based agents in various scenarios. 
We also attempt to extend our work to real-world multi-agent applications, to verify it can scale up the number of agents in building multi-agent scenarios.

\subsection{\method for MetaGPT}
MetaGPT~\citep{hong2024metagpt} is a meta-programming framework that enhances LLM-based multi-agent collaborations by integrating efficient human workflows. 
It employs an assembly line approach to assign diverse roles to agents, effectively simplifying complex tasks into manageable subtasks that multiple agents can execute collaboratively. 
As shown in Figure~\ref{figue:metagpt}, we choose the debate scenario used in MetaGPT, which includes two debaters with different opinions, leading to dull and repetitive content generation. 
Here, instead of manually assigning new roles, we applied \method to extend each debate team to more agents with diverse settings, increasing the variety of opinions and the quality of the debate.

\subsection{\method for Camel}
Camel~\citep{li2023camel} is recognized for its framework that supports communicative role-playing agents. 
Initially, humans establish this framework by conceptualizing an idea and designing specific roles, such as the AI assistant role and the AI user role. 
These roles are then assigned to the assistant and user agents, respectively, enabling them to fulfill the task. 
As illustrated in Figure~\ref{figue:001}, \method can be utilized to automatically produce agents from AI assistants for interaction with AI users, bypassing the need for manual role design.

\subsection{\method for AutoGen}
\begin{figure*}[t]
    \centering
    \includegraphics[width=0.8\linewidth]{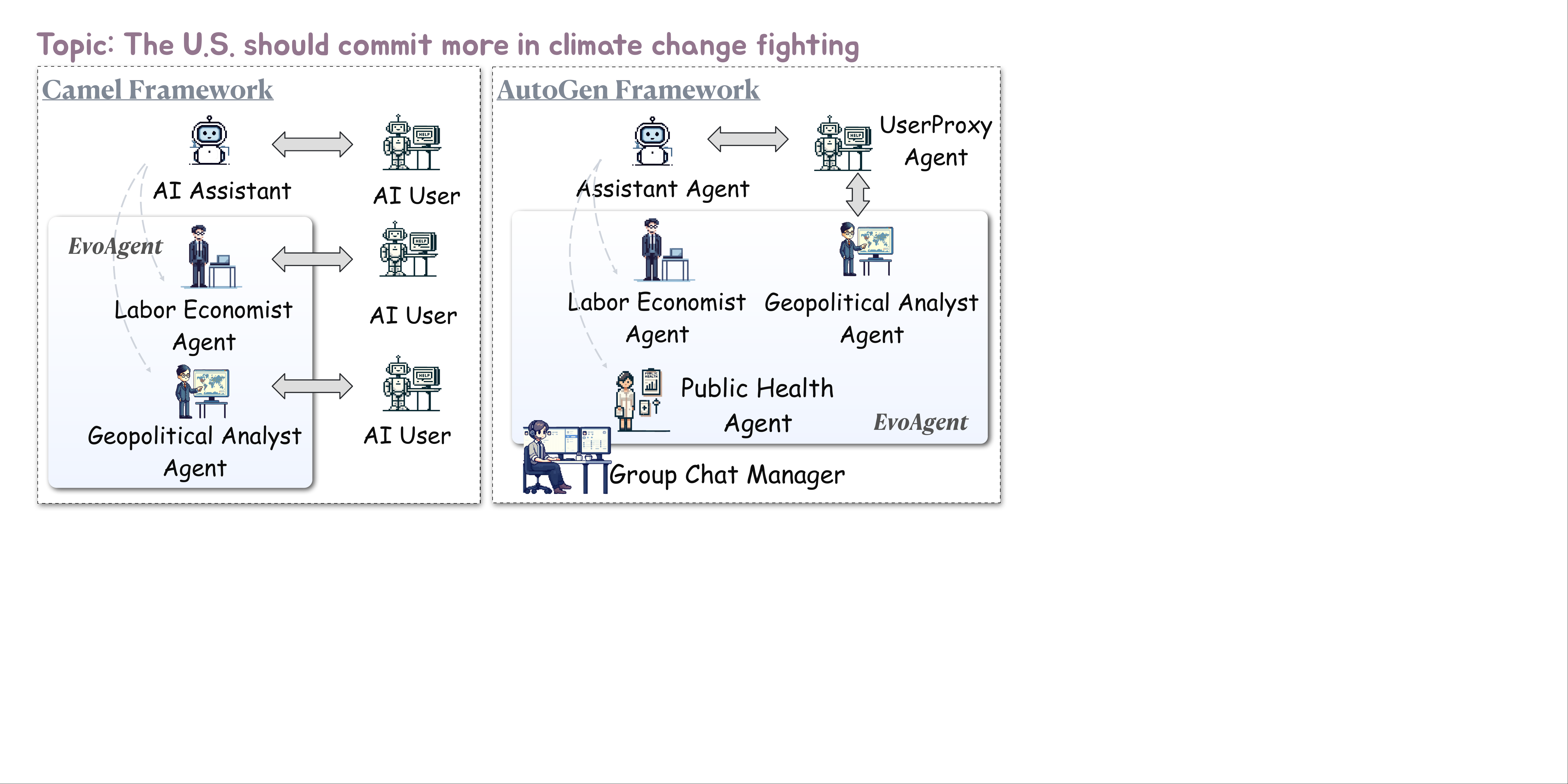}
    \caption{The adaption of \method on Camel and AutoGen frameworks.}
    \label{figue:001}
\end{figure*}

AutoGen~\citep{wu2023autogen} offers a framework that enables the creation of customizable and conversable agents by integrating various LLMs.
Initially, humans configure the assistant agents along with a user proxy agent. 
Then, a group chat manager is responsible for selecting a speaker, gathering responses, and disseminating the message. 
As depicted in Figure~\ref{figue:001}, \method facilitates the creation of multiple expert roles from a single assistant agent, thereby increasing the agent number in group chats without the need for manual design.

\begin{figure*}[t]
\centering
\begin{minipage}{\textwidth}
\lstset{
    backgroundcolor=\color[RGB]{245,245,244},
    breaklines=true,
    breakindent=0pt,
    basicstyle=\ttfamily\small,
    emph={Direct,CoT,Self,Method,Feedback,Multi,agents,Generation,Result,Refine,OverGen,Quality,Check},
    emphstyle={\bfseries\color{NavyBlue}}
}\begin{lstlisting}[caption={Instruction templates for for 0-shot learning (Direct), Chain-of-thought (CoT) prompting and Self-Refine method},label=lst:self-refine]

Direct Method:
{question}
Answer:

CoT Method:
{question}
You need to give reasons first and then give the answer.
Answer:

Self-Refine Method:

Step One: Feedback Generation:
You are a helpful assistant that provides feedback on {task}

{question}
This is the answer from a student: {answer}.

Please do not refine the answer but give some insightful suggestions for the student to help him better answer the question.
Suggestion:

Step Two: Result Refine:
{question}
This is your answer:
{answer}

There is the suggestion from an assistant:
Suggestion: {feedback}

Now you can refine your answer with his suggestion to better answer the question. 
Keep in mind that his suggestion may not be correct, so critically decide whether to accept his response or stick with your original one.
You need to give reasons first and then give the answer.
Revised Answer:


\end{lstlisting}
\end{minipage}\hfill
\begin{minipage}{\textwidth}
\end{minipage}
\end{figure*}

\begin{figure*}[t]
\centering
\begin{minipage}{\textwidth}
\lstset{
    backgroundcolor=\color[RGB]{245,245,244},
    breaklines=true,
    breakindent=0pt,
    basicstyle=\ttfamily\small,
    emph={Direct,CoT,Self,Method,Feedback,Multi,agents,Generation,Result,Refine,OverGen,Quality,Check},
    emphstyle={\bfseries\color{NavyBlue}}
}\begin{lstlisting}[caption={Instruction templates for SPP method},label=lst:ssp]
When faced with a task, begin by identifying the participants who will contribute to solving the task. Then, initiate a multi-round collaboration process until a final solution is reached. The participants will give critical comments and detailed suggestions whenever necessary.

Here are some examples:
---
Example Task 1: Use numbers and basic arithmetic operations (+ - * /) to obtain 24. You need to use all numbers, and each number can only be used once.
Input: 6 12 1 1
Participants: AI Assistant (you); Math Expert
Start collaboration!
Math Expert: Let's analyze the task in detail. You need to make sure that you meet the requirement, that you need to use exactly the four numbers (6 12 1 1) to construct 24. To reach 24, you can think of the common divisors of 24 such as 4, 6, 8, 3 and try to construct these first. Also you need to think of potential additions that can reach 24, such as 12 + 12.
AI Assistant (you): Thanks for the hints! Here's one initial solution: (12 / (1 + 1)) * 6 = 24
Math Expert: Let's check the answer step by step. (1+1) = 2, (12 / 2) = 6, 6 * 6 = 36 which is not 24! The answer is not correct. Can you fix this by considering other combinations? Please do not make similar mistakes.
AI Assistant (you): Thanks for pointing out the mistake. Here is a revised solution considering 24 can also be reached by 3 * 8: (6 + 1 + 1) * (12 / 4) = 24.
Math Expert: Let's first check if the calculation is correct. (6 + 1 + 1) = 8, 12 / 4 = 3, 8 * 3 = 24. The calculation is correct, but you used 6 1 1 12 4 which is not the same as the input 6 12 1 1. Can you avoid using a number that is not part of the input?
AI Assistant (you): You are right, here is a revised solution considering 24 can be reached by 12 + 12 and without using any additional numbers: 6 * (1 - 1) + 12 = 24.
Math Expert: Let's check the answer again. 1 - 1 = 0, 6 * 0 = 0, 0 + 12 = 12. I believe you are very close, here is a hint: try to change the "1 - 1" to "1 + 1".
AI Assistant (you): Sure, here is the corrected answer:  6 * (1+1) + 12 = 24
Math Expert: Let's verify the solution. 1 + 1 = 2, 6 * 2 = 12, 12 + 12 = 12. You used 1 1 6 12 which is identical to the input 6 12 1 1. Everything looks good!
Finish collaboration!
Final answer: 6 * (1 + 1) + 12 = 24

---
{question}
\end{lstlisting}
\end{minipage}\hfill
\begin{minipage}{\textwidth}
\end{minipage}
\end{figure*}

\begin{figure*}[t]
\centering
\begin{minipage}{\textwidth}
\lstset{
    backgroundcolor=\color[RGB]{245,245,244},
    breaklines=true,
    breakindent=0pt,
    basicstyle=\ttfamily\small,
    emph={Multi,Generation,Result,Refine,OverGen,Quality,Check,Crossover,Mutation,Update},
    emphstyle={\bfseries\color{NavyBlue}}
}\begin{lstlisting}[caption={Instruction templates for \method},label=lst:instruction]

Crossover and Mutation:

{question}

This is your result:
{answer}

Now, you can create and collaborate with multiple experts to improve your result. Therefore, please describe in as much detail as possible the different skills and focuses you need from multiple experts individually. We will provide each expert with the same information and query. However, please note that each profession has its own specialization, so you can assign each expert to just one sub-task to ensure a more refined response. We will relay their responses to you in turn, allowing you to reorganize them into a better answer. Please note that the description should be narrated in the second person, for example: You are a XXX.

These are the descriptions of the experts you have created before for this task:
{description}

Therefore, you need to follow two principles:
1. Crossover: You need to check which skills should be improved in the previous agents and then update them.
2. Mutation: You need to make new agents distinct from previous agents while maintaining their task-solving capability.

Now, you can give the description for a new expert (Please note that only be one, do not give multiple at one time):


Quality Check:

{question}

We employ mulitple experts to answer this query. The following is a second-person introduction to the experts we have hired:
{description_ls}

Now, we will hire a new expert to help better respond to user query. Here is a second person description of the new expert: {description}
Please evaluate the new expert based on the following criteria to decide whether they should be retained or not:
1. The new expert is distinct and does not duplicate any previously hired experts.
2. Based on the new expert's description, determine if they can effectively assist in answering users' questions.
Give the reason first and then give the choice. If retaining, please reply with: Retain. If discarding, please reply with: Discard.

Result Update:

{question}
This is your result:
{old_answer}

You invite an expert whose description is: {description}
This expert also give his answer based on his own professional knowledge: {new_answer}.

Now you can refine your result with his answer to better answer the question. 
Keep in mind that his answer may not be correct, so critically decide whether to accept his response or stick with your original one.
Revised Answer:

\end{lstlisting}
\end{minipage}\hfill
\begin{minipage}{\textwidth}
\end{minipage}
\end{figure*}

\begin{figure*}[t]
\centering
\begin{minipage}{\textwidth}
\lstset{
    backgroundcolor=\color[RGB]{245,245,244},
    breaklines=true,
    breakindent=0pt,
    basicstyle=\ttfamily\small,
    emph={OverGen,Selection,All,in,PK,Suggest,PromptRefine},
    emphstyle={\bfseries\color{NavyBlue}}
}\begin{lstlisting}[caption={Instruction templates of different ablation settings in \method.},label=lst:ablation]


PK:
{question}
We invite {n} experts. They give the results based on their own professional knowledge:
Here are second-person descriptions of these experts with their answers:
{select}
Now you can should help us select the best result which can meet the query. 
You need to give reasons first and then give the answer with the format: "Final Answer: Expert #XX"

All-in:
{question}
This is your answer: {old_answer}.
Furthermore, you also invite {n} experts. They also give answers based on their own professional knowledge:
Here are second person descriptions of these experts with their answers:
{description_ls}
Now you can refine your answer with these answers to better meet the query.

Suggest:
{specialized_Agent_description}
{question}
This is the result from an AI assistant: {answer}.
Please do not refine the plan but give some insightful suggestions for the AI assistant to help it better meet the user's query.
Suggestion:

OverGen:
{question}
Please generate three different results at one time for user to choose from. The format can be:
Result #1:
Result #2:
Result #3:
Three Different Candidate Results:

PromptRefine:
{question}
This is the result from an AI assistant, which description is 
"{original_description}": 
{answer}.
Please do not refine the result but refine the description of the AI assistant to help it better answer the user's query.
Please note that the description should be narrated in the second person, for example: You are a XXX.
Description:


\end{lstlisting}
\end{minipage}\hfill
\begin{minipage}{\textwidth}
\end{minipage}
\end{figure*}

\begin{figure*}[t]
\centering
\begin{minipage}{\textwidth}
\lstset{
    backgroundcolor=\color[RGB]{245,245,244},
    breaklines=true,
    breakindent=0pt,
    basicstyle=\ttfamily\small,
    emph={CoT,EvoAgent,Specialized,Answer,Reason,Expert,Agent},
    emphstyle={\bfseries\color{NavyBlue}}
}\begin{lstlisting}[caption={An example of multi-agent generation from \method for Logic Grid Puzzle},label=lst:example_logic]
Question: There are 5 houses in a row, numbered 1 on the left to 5 on the right. There is one person living in each house. The people in these houses have different characteristics:
- Each person has a different kind of car: one owns an SUV, one owns a sedan, one drives a convertible, one drives a minivan, and one drives a truck
- Each person has a favorite color: one likes red, one likes yellow, one likes green, one likes white, and one likes blue
- Each person is wearing different shoes: one is wearing running shoes, one is wearing boots, one is wearing high heels, one is wearing sandals, and one is wearing loafers
Clue(s):
1. There is one house between where the person who is wearing loafers lives and where the person who drives a minivan lives. 2. The person who drives a truck does not live in the fourth house. 3. The person who drives a truck lives directly left of the person who is wearing sandals. 4. The person who likes red lives somewhere to the right of the person who likes yellow. 5. The person who is wearing loafers lives somewhere to the left of the person who drives a truck. 6. The person who likes blue lives in the fifth house. 7. There are two houses between where the person who owns a sedan lives and where the person who is wearing running shoes lives. 8. The person who is wearing boots lives somewhere to the left of the person who drives a truck. 9. The person who likes white lives in the fourth house. 10. There is one house between where the person who likes green lives and where the person who drives a truck lives. 11. The person who owns an SUV does not live in the first house.
What is the number of the house where the person who is wearing boots lives? choice: 2, choice: 5, choice: 3, choice: 1, choice: 4.

EvoAgent:
Specialized Agent #1 Description:
You are a logician, specializing in logical reasoning and problem-solving. Your task is to analyze the given clues and use deductive reasoning to determine the correct sequence of houses, their occupants, and their characteristics. You will need to consider each clue carefully, understanding how it relates to the others and how it narrows down the possibilities. Your focus should be on ensuring the logical consistency of the solution, making sure that all clues are satisfied without contradiction.
Reason: From clue 6 and 9, we know that the person who likes blue lives in the fifth house and the person who likes white lives in the fourth house. From clue 4, the person who likes red lives somewhere to the right of the person who likes yellow. Since the fourth and fifth houses are occupied by the people who like white and blue, the person who likes red must live in the third house and the person who likes yellow must live in the first or second house.From clue 10, there is one house between where the person who likes green lives and where the person who drives a truck lives. Since the third, fourth, and fifth houses are occupied by the people who like red, white, and blue, the person who likes green must live in the first or second house. Therefore, the person who likes yellow must live in the first house and the person who likes green must live in the second house.From clue 11, the person who owns an SUV does not live in the first house. So, the person who owns an SUV must live in the second, third, or fourth house.From clue 2, the person who drives a truck does not live in the fourth house. So, the person who drives a truck must live in the second or third house.From clue 3, the person who drives a truck lives directly left of the person who is wearing sandals. So, the person who drives a truck must live in the second house and the person who is wearing sandals must live in the third house.From clue 8, the person who is wearing boots lives somewhere to the left of the person who drives a truck. So, the person who is wearing boots must live in the first house.
Answer: choice: 1

\end{lstlisting}
\end{minipage}\hfill
\begin{minipage}{\textwidth}
\end{minipage}
\end{figure*}

\begin{figure*}[t]
\centering
\begin{minipage}{\textwidth}
\lstset{
    backgroundcolor=\color[RGB]{245,245,244},
    breaklines=true,
    breakindent=0pt,
    basicstyle=\ttfamily\small,
    emph={Specialized,CoT,EvoAgent,Answer,Reason,Expert,Plan,Story,Evaluation,Agent},
    emphstyle={\bfseries\color{NavyBlue}}
}\begin{lstlisting}[caption={An example of multi-agent generation from \method for Trivia Creative Writing},label=lst:example_tcw]
Write a short and coherent story about Zootopia that incorporates the answers to the following 5 questions:
1. Following success at the 1979 election whose party conference speeech included the lines 'you turn if you want to, the lady's not for turning'?
2. Which snooker player was simply known as 'The Grinder'?
3. Which former West Indian fast bowler was known as 'Whispering Death'?
4. Which Eastenders character had a dog called Willy?
5. In which year did the first episode of Minder take place?

EvoAgent:
Specialized Agent #1 Description:
You are a sports historian with a deep understanding of snooker and cricket from the 1970s and 1980s. You are familiar with the careers and playing styles of Cliff Thorburn, known as 'The Grinder', and Michael Holding, known as 'Whispering Death'. Your role in this task will be to provide context and insight into these sports figures and their significance in the story about Zootopia.

Specialized Agent #2 Description:
You are a political historian with a deep understanding of the political climate and key figures of the 1970s and 1980s, particularly in the United Kingdom. You have extensive knowledge about Margaret Thatcher, her policies, and her impact on society. Your role in this task will be to provide context and insight into the significance of the reference to Thatcher's famous speech in the story about Zootopia.

Story: In the vibrant city of Zootopia, the year was 1979, a time of significant change and excitement. The city had just elected its first female mayor, a lioness named Margaret Thatcher, known for her unwavering determination and strong leadership. Her victory speech at the party conference echoed the famous line from the human world's Margaret Thatcher, "You turn if you want to, the lady's not for turning." This phrase resonated with the citizens of Zootopia, who admired her steadfastness and determination. In the heart of the city, a meticulous badger known as 'The Grinder' ran a popular snooker club. His nickname was a tribute to the legendary human snooker player Cliff Thorburn, known for his relentless and meticulous style of play. The Grinder's snooker club was a hub of social activity, where animals from all walks of life came to unwind and engage in friendly competition. Meanwhile, a new recruit had joined the Zootopia Police Department, a cheetah named Michael Holding. Known as 'Whispering Death' during his cricketing days, a moniker borrowed from the former West Indian fast bowler, he was now a respected figure in the community, using his speed and stealth to uphold the law. In another part of the city, a popular soap opera was being filmed. The show was called "Eastenders of Zootopia," and one of the beloved characters was a rabbit named Ethel Skinner, who had a dog named Willy. This was a charming reference to the human Eastenders character of the same name, symbolizing the enduring friendships that were a cornerstone of Zootopian society. The year 1979 was also significant as it marked the airing of the first episode of the Zootopian version of the human show "Minder". The show was a reflection of the city's vibrant and diverse culture, much like the original British series. It was a testament to the city's ability to adapt and evolve, while still holding onto its rich history and traditions. In conclusion, Zootopia was a melting pot of cultures and histories, much like the television shows and characters it referenced. These references served as a bridge between the past and the present, reminding the citizens of their roots while encouraging them to embrace the future.

Evaluation: correct_count=5/5
\end{lstlisting}
\end{minipage}\hfill
\begin{minipage}{\textwidth}
\end{minipage}
\end{figure*}

\begin{figure*}[t]
\centering
\begin{minipage}{\textwidth}
\lstset{
    backgroundcolor=\color[RGB]{245,245,244},
    breaklines=true,
    breakindent=0pt,
    basicstyle=\ttfamily\small,
    emph={Specialized,CoT,EvoAgent,Answer,Reason,Expert,Final,Agent},
    emphstyle={\bfseries\color{NavyBlue}}
}\begin{lstlisting}[caption={An example of multi-agent generation from \method for Codenames Collaborative},label=lst:example_code]
Try to identify the 3 words best associated with the word "fruit" from the following word list: 
["bowl", "einstein", "banana", "brain", "judge", "groom", "crusader", "stamp", "peach", "halloween", "troll", "slipper", "earthquake", "tip", "sun", "quarter", "pad", "nude", "helmet", "laundry", "wedding", "mirror", "sleep", "minotaur", "wish"].

EvoAgent:
Specialized Agent #1 Description:
You are a cultural anthropologist with a focus on symbolism and rituals. Your expertise in understanding how different objects, including fruits, are used and symbolized in various cultures will be beneficial in this task. You can provide insights into how certain words might be associated with "fruit" from a cultural or symbolic perspective. Your focus will be on the cultural and symbolic aspects of the task.

Reason:
1. 'Banana': This is a type of fruit, rich in potassium and dietary fiber. It is commonly consumed raw and is used in a variety of dishes like desserts, smoothies, and baked goods.
2. 'Peach': This is also a type of fruit, known for its sweet and juicy nature. Peaches are rich in vitamins A and C and are often eaten raw or used in dishes like pies, jams, and salads.
3. 'Bowl': This word is associated with fruit in the context of a fruit bowl, which is a common way to store and display various types of fruit. A bowl of fruit can also be a healthy snack or dessert option.
Answer: Banana, Peach, Bowl

Specialized Agent #2 Description:
You are a nutritionist with a deep understanding of different types of food, including fruits. Your knowledge about the nutritional value, common uses, and cultural significance of various fruits will be helpful in this task. You can provide insights into how certain words might be associated with "fruit" from a dietary or nutritional perspective. Your focus will be on the nutritional and dietary aspects of the task.

Reason:
1. 'Banana': This is a direct association as a banana is a type of fruit. In many cultures, bananas symbolize fertility and prosperity. In Hinduism, for example, bananas are considered a sacred fruit and are used in religious rituals.
2. 'Peach': Similar to banana, a peach is also a type of fruit. In Chinese culture, peaches are symbols of immortality and unity. They are often used in traditional rituals and ceremonies.
3. 'Bowl': A bowl is often used to hold fruits. In many cultures, a bowl of fruit is a symbol of abundance, fertility, and prosperity. It is often used in rituals and ceremonies as an offering or a centerpiece.
Answer: banana, peach, bowl

Final Answer:
After considering the expert's input, I find that their reasoning aligns with my initial thoughts. The words 'banana' and 'peach' are indeed types of fruit, and their nutritional value and common uses further strengthen their association with the word 'fruit'. The word 'bowl' is also associated with fruit in the context of a fruit bowl, which is a common way to store and display various types of fruit. Therefore, I agree with the expert's choices.
Final Answer: banana, peach, bowl.
\end{lstlisting}
\end{minipage}\hfill
\begin{minipage}{\textwidth}
\end{minipage}
\end{figure*}

\begin{figure*}[t]
\centering
\begin{minipage}{\textwidth}
\lstset{
    backgroundcolor=\color[RGB]{245,245,244},
    breaklines=true,
    breakindent=0pt,
    basicstyle=\ttfamily\small,
    emph={Specialized,CoT,EvoAgent,Answer,Reason,Expert,Agent},
    emphstyle={\bfseries\color{NavyBlue}}
}\begin{lstlisting}[caption={An example of multi-agent generation from \method for MMMU dataset.},label=lst:example_mmmux]
Question: <image 1> The elliptical orbit of a planet around the Sun is shown on the diagram. Which of the following statements is true?
A. the eccentricity of the orbit is less than zero
B. the eccentricity of the orbit is greater than 1
C. the sun might be at point C
D. the sun might be at point D
E. the sun might be at point B (Correct)

CoT:
Reason: In an elliptical orbit, the Sun is located at one of the two foci of the ellipse. The eccentricity of an orbit is a measure of how much the orbit deviates from a perfect circle, with 0 being a perfect circle and 1 being a parabola. Since the orbit shown is an ellipse, the eccentricity must be between 0 and 1. Points A and E are the farthest points from the center of the ellipse, so the Sun cannot be located at either of those points. Point B is at the center of the ellipse, so the Sun cannot be located there either. This leaves points C and D as possible locations for the Sun. However, since the Sun must be at one of the foci and point D is closer to the center of the ellipse than point C, it is more likely that the Sun is located at point C.
Answer: C. the sun might be at point C

EvoAgent:
Specialized Agent #1 Description:
You are a mathematician with a specialization in geometry, particularly in the study of conic sections which include ellipses. Your expertise allows you to understand the properties and definitions of ellipses, including the concept of foci. You can provide a mathematical perspective on why the Sun must be located at one of the foci of the elliptical orbit and not at any other point. Your knowledge will help explain the geometric reasons behind the positioning of the Sun in relation to the planet's orbit, and why the given options in the statements do not align with the mathematical definition of an ellipse.
Reason: An ellipse is defined as the set of all points for which the sum of the distances to two fixed points, called foci, is constant. The Sun, being one of the massive bodies in the solar system, must be located at one of the foci of the elliptical orbit of a planet due to the gravitational forces that govern the motion of the planets. Option E is correct because the Sun might be at point B, which is one of the foci of the ellipse. The other focus would be at a point symmetrically opposite to B along the major axis.
Answer: E. the sun might be at point B.

Specialized Agent #2 Description:
You are an astronomer with a specialization in observational astronomy. Your expertise lies in the practical observation and data collection of celestial bodies, including the use of telescopes and other instruments to study the positions and movements of planets and stars. You have a deep understanding of how the apparent motion of celestial bodies can be used to infer their actual motion in space. You can provide insight into how observations of planetary orbits, such as the one depicted in the diagram, have historically been used to deduce the laws of planetary motion and the positions of stars like the Sun. Your knowledge will help explain how practical observations support the theoretical understanding that the Sun must be located at one of the foci of the elliptical orbit.
Reason: The eccentricity of an orbit is a measure of how much it deviates from being a perfect circle. It ranges from 0 (a perfect circle) to 1 (a parabola). An eccentricity less than zero or greater than 1 is not possible for an elliptical orbit. Therefore, options A and B are incorrect. According to Kepler's laws of planetary motion, the Sun must be located at one of the foci of the elliptical orbit. Point B is located at a position that could be one of the foci of the ellipse.
Answer: E. the sun might be at point B.

\end{lstlisting}
\end{minipage}\hfill
\begin{minipage}{\textwidth}
\end{minipage}
\end{figure*}

\begin{figure*}[t]
\centering
\begin{minipage}{\textwidth}
\lstset{
    backgroundcolor=\color[RGB]{245,245,244},
    breaklines=true,
    breakindent=0pt,
    basicstyle=\ttfamily\small,
    emph={Specialized,CoT,EvoAgent,Answer,Reason,Expert,Task,Agent},
    emphstyle={\bfseries\color{NavyBlue}}
}\begin{lstlisting}[caption={An example of multi-agent generation from \method for ScienceWorld.},label=lst:scienceworld]
Task 1: Your task is to determine if a metal fork is electrically conductive. The metal fork is located around the kitchen. First, focus on the metal fork. If conductive, place it in the blue box. If nonconductive, place it in the green box.
EvoAgent:
Specialized Agent Description:
You are a thermodynamics expert. Your expertise in heat transfer and temperature measurement is crucial for this task. You understand how to properly use a thermometer and interpret its readings. You also have knowledge about the properties of various substances, including their typical temperature ranges. Your task is to guide the assistant in using the thermometer to accurately measure the temperature of the unknown substance B.

Task 2: Your task is to boil gallium. For compounds without a boiling point, combusting the substance is also acceptable. First, focus on the substance. Then, take actions that will cause it to change its state of matter.
EvoAgent:
Specialized Agent Description:
You are a metallurgist with a deep understanding of the properties and behaviors of metals, including gallium. Your expertise includes knowledge of the melting and boiling points of various metals, as well as the processes and equipment needed to heat them to these temperatures. You are also familiar with safety procedures and precautions when handling and heating metals. Your task is to provide detailed instructions on how to safely and effectively boil gallium in the foundry.

Task 3: Your task is to change the state of matter of gallium. First, focus on the substance. Then, take actions that will cause it to change its state of matter.
EvoAgent:
Specialized Agent Description:
You are a chemist with a specialization in the properties and behaviors of substances. You have a deep understanding of the state changes of matter, specifically how substances like gallium can transition from solid to liquid or gas. You are familiar with the conditions required for these changes to occur, such as temperature and pressure. You also have knowledge of the equipment and procedures used in a foundry, which will be useful in this task. Your role is to provide guidance on how to manipulate the conditions in the foundry to achieve the desired state change in gallium.


\end{lstlisting}
\end{minipage}\hfill
\begin{minipage}{\textwidth}
\end{minipage}
\end{figure*}

\begin{figure*}[t]
\centering
\begin{minipage}{\textwidth}
\lstset{
    backgroundcolor=\color[RGB]{245,245,244},
    breaklines=true,
    breakindent=0pt,
    basicstyle=\ttfamily\small,
    emph={Specialized,CoT,EvoAgent,Answer,Reason,Expert,Agent},
    emphstyle={\bfseries\color{NavyBlue}}
}\begin{lstlisting}[caption={An example of multi-agent generation from \method for TravelPlanner dataset.},label=lst:example_travelplanner]
Question: Can you help me with a travel plan departing from Buffalo to Atlanta for a duration of 3 days, specifically from March 2nd to March 4th, 2022? I plan to travel alone and my planned budget for the trip is around $1,100.

EvoAgent:
Specialized Agent #1 Description:
You are a culinary expert with a deep understanding of food culture in Atlanta. Your expertise includes knowledge of the best local restaurants, popular dishes, and dietary restrictions. You can provide recommendations for breakfast, lunch, and dinner spots that offer a unique taste of Atlanta's food scene. Your suggestions should consider the traveler's budget and the proximity of the restaurants to the attractions and accommodation listed in the plan.

Specialized Agent #2 Description:
You are a travel logistics expert with extensive knowledge of transportation options and costs between Buffalo and Atlanta. Your expertise includes understanding flight schedules, costs, and durations, as well as alternative transportation methods such as self-driving and taxi services. You can provide the most cost-effective and time-efficient transportation options for the traveler, considering their budget and the travel dates provided in the plan.

Specialized Agent #3 Description:
You are a local tourism expert with comprehensive knowledge of Atlanta's attractions. Your expertise includes understanding the city's popular tourist spots, historical sites, museums, parks, and unique local experiences. You can provide recommendations for attractions that the traveler can visit during their stay, considering their budget and the proximity of these attractions to the traveler's accommodation and meal spots. Your suggestions should also take into account the operating hours of these attractions to ensure they align with the traveler's schedule.

Answer:
Day 1:
Current City: from Buffalo to Atlanta
Transportation: Flight Number: F3502691, from Buffalo to Atlanta, Departure Time: 18:48, Arrival Time: 20:51
Breakfast: Not applicable as the flight is in the evening.
Attraction: Not applicable as the flight is in the evening.
Lunch: Not applicable as the flight is in the evening.
Dinner: Saffron, Atlanta
ccommodation: Spacious private room close St. Barnabas Hospital, Atlanta
Day 2:
Current City: Atlanta
Transportation: -
Breakfast: Adda, Atlanta
Attraction: Atlanta Botanical Garden, Atlanta; World of Coca-Cola, Atlanta.
Lunch: Baba Au Rhum, Atlanta
Dinner: Asian Bistro, Atlanta
Accommodation: Spacious private room close St. Barnabas Hospital, Atlanta
Day 3:
Current City: from Atlanta to Buffalo
Transportation: Flight Number: F3500648, from Atlanta to Buffalo, Departure Time: 21:24, Arrival Time: 23:26
Breakfast: Chef Style, Atlanta
Attraction: Georgia Aquarium, Atlanta; Martin Luther King, Jr. National Historical Park, Atlanta.
Lunch: Pizza Central, Atlanta
Dinner: Daawat-e-Kashmir, Atlanta
Accommodation: -.

\end{lstlisting}
\end{minipage}\hfill
\begin{minipage}{\textwidth}
\end{minipage}
\end{figure*}

\end{appendix}

\end{document}